\documentclass{article}
\usepackage{latexsym}
\usepackage{amsmath,amsthm,amsopn,amssymb}
\usepackage{bbding}
\usepackage{graphicx}
\usepackage{hyperref}
\usepackage{mathptmx}
\allowdisplaybreaks

\numberwithin{equation}{section}
\newtheorem{theorem}{Theorem}[section]
\newtheorem{lemma}{Lemma}[section]

\newtheorem{algorithm}{Algorithm}[section]

\makeatother

\begin{document}

\title{regularization  for convolutional kernel tensors  to avoid unstable gradient problem in convolutional neural networks
%\thanks{This research is supported by the Fundamental Research Funds for the Central Universities in China University of Geosciences, Beijing (2652017140).}
}

\author{
%Authors
Pei-Chang Guo \thanks{ e-mail:peichang@cugb.edu.cn} \\
School of Science,
China University of Geosciences, Beijing, 100083, China\\
}

\date{}

\maketitle
\begin{abstract}
Convolutional neural networks are very popular nowadays. Training neural networks is not an easy task. Each convolution corresponds to a structured transformation matrix. In order to help avoid the exploding/vanishing gradient problem, it is desirable that the singular values of each transformation matrix are not large/small in the training process.  We propose three new  regularization terms for a convolutional kernel tensor to constrain the singular values of each transformation matrix.   We show how to carry out the gradient type methods, which provides  new insight about  the training of convolutional neural networks.

\vspace{2mm} \noindent \textbf{Keywords}:  regularization, singular values, doubly  block banded Toeplitz matrices, convolution, tensor.
\end{abstract}

\section{Introduction}

As we know, each convolution arithmetic corresponds to a linear structured transformation matrix. We use $vec(X)$ to denote the vectorization of $X$. If $X$ is a matrix,  $vec(X)$ is the column vector got  by  stacking the columns of  $X$ on top of one another. If $X$ is a tensor,  $vec(X)$ is the column vector got  by  stacking the columns of the flattening of $X$ along the first index (see \cite{golub2012} for more on flattening of a tensor). We use $*$ to denote the convolution arithmetic in deep learning.  Given a kernel $K$, the output $Y=K*X$ can be reshaped through
 \begin{equation*}
    vec(Y)=Mvec(X),
\end{equation*}
where $M$ is the  linear transformation matrix.

When training the deep neural networks, gradient exploding and vanishing are fundamental obstacles. It's helpful to make the largest singular value of $M$ be smaller for controlling exploding gradients and it's helpful to make the smallest singular value of $M$ be larger for controlling vanishing gradients. In this paper we will give three regularization terms about convolutional kernel $K$ to change the singular values of $M$ and show how to carry out gradient type methods for them.

  When we refer to convolution in deep learning, there is no flip operation and only element-wise multiplication and addition are performed.  Besides, in the field of deep learning, depending on different strides and padding patterns, there are many different forms of convolution arithmetic\cite{dumoulin2018}. Without losing generality, in this paper we will adopt the same convolution with unit strides.   We use $\ulcorner \cdot\urcorner$ is to round a number to the nearest integer greater than or equal to that number. If a convolutional kernel is a matrix $K\in \mathbb{R}^{k\times k}$  and the input is a matrix  $X\in \mathbb{R}^{N\times N}$,  each entry of the output $Y\in \mathbb{R}^{N \times N}$  is produced by
\begin{equation*}
    Y_{r,s}= (K*X)_{r,s}= \sum_{p\in \{1,\cdots,k\} }\sum_{q\in \{1,\cdots,k\}} X_{r-m+p,s-m+q}K_{p,q},
\end{equation*}
where $m=\ulcorner k/2\urcorner$, , and $X_{i,j}=0$ if $i\leq 0$ or $i> N$, or $j\leq 0$ or $j> N$.

In convolutional neural networks, usually there are multi-channels and a convolutional kernel is  represented by a 4 dimensional tensor.
If a convolutional kernel is a 4 dimensional tensor $K\in \mathbb{R}^{k\times k\times g  \times h}$  and the input is 3 dimensional tensor $X\in \mathbb{R}^{N\times N\times g}$, each entry of  the output $Y\in \mathbb{R}^{N \times N \times h}$  is produced by
\begin{equation*}
    Y_{r,s,c}= (K*X)_{r,s,c}=\sum_{d\in \{1,\cdots,g\}} \sum_{p\in \{1,\cdots,k\} }\sum_{q\in \{1,\cdots,k\}} X_{r-m+p,s-m+q,d}K_{p,q,d,c},
\end{equation*}
where $m=\ulcorner k/2\urcorner$ and $X_{i,j,d}=0$ if $i\leq 0$ or $i> N$, or $j\leq 0$ or $j> N$.

In the community of deep learning, there have been papers devoted to enforcing  the orthogonality or spectral norm regularization on the weights of a neural network \cite{brock2017,cisse,miyato2018,yoshida}. The difference between our paper and papers including \cite{brock2017,cisse,miyato2018,yoshida} and the references therein  is about how to handle convolutions. They enforce the constraint directly on the $h\times (gkk)$  matrix reshaped from the kernel $K\in \mathbb{R}^{k\times k\times g  \times h}$,  while we enforce the the  constraint on the transformation matrix $M$ corresponding to the convolution kernel $K$. In  \cite{sedghi2018}, the authors project a convolutional  layer onto the set of layers obeying a bound on the operator norm of the layer and use numerical results to show this is an effective regularizer. A drawback of the method in \cite{sedghi2018} is that projection can prevent the singular values of the transformation matrix being large but can't avoid the singular values to be too small.

In \cite{guo2019,guo20192,Orthberkely}, regularization methods are proposed  to let  the corresponding transformation matrices be orthogonal, where the approach is to minimize the norm of $M^TM-I$. In this paper we propose new regularization methods for the convolutional kernel tensor $K$, which can reduce the largest singular value  and increase the smallest singular value of  $M$ independently or simultaneously depending on the need in the training process.

The rest of the paper is organized as follows.
As we have mentioned, the input channels and the output channels maybe more than one so the kernel is usually represented by a tensor $K\in \mathbb{R}^{k\times k\times g  \times h}$.
In Section~\ref{sec:one},   we propose the penalty functions and calculate the partial derivatives  for the case that the kernel $K$ is a $k\times k$ matrix. In Section \ref{sec:multi}, we propose the penalty functions and calculate the partial derivatives for the case that $K$ is a $k\times k\times g  \times h$ tensor.
In Section~\ref{sec:numer},  we present numerical results to show the method is feasible and effective. In Section~\ref{sec:conclu}, we will give some conclusions and discuss some work that may be done in the future.

\section{penalty function for one-channel convolution}\label{sec:one}

When the numbers of input channels and the output channels are both $1$, the convolutional kernel are a $k\times k$ matrix.
Assuming the data matrix is $N\times N$, we use a $3\times 3$ matrix as a convolution kernel to show the associated structured transformation matrix. Let $K$ be the convolutional kernel,
\begin{eqnarray*}
K=\left(\begin{array}{ccc}
k_{11} & k_{12} & k_{13} \\
k_{21} & k_{22} & k_{23} \\
k_{31} & k_{32} & k_{33}
\end{array}\right).
\end{eqnarray*}
Then the transformation matrix $M$ such that $vec(Y)=Mvec(X)$ for $Y=K*X$ is
\begin{eqnarray}\label{conv0}
M=\left(
  \begin{array}{cccccc}
    A_0 & A_{-1} & 0 & 0 & \cdots & 0 \\
    A_1 & A_0 & A_{-1} & \ddots & \ddots & \vdots \\
    0 & A_1 & A_0 & \ddots & \ddots & 0 \\
    0 & \ddots & \ddots & \ddots & A_{-1} & 0 \\
    \vdots & \ddots & \ddots & A_1 & A_0 & A_{-1} \\
    0 & \cdots & 0 & 0 & A_1 & A_0 \\
  \end{array}
\right)
\end{eqnarray}
where
\begin{eqnarray*}
A_0=\left(
  \begin{array}{cccccc}
    k_{22} & k_{32} & 0 & 0 & \cdots & 0 \\
    k_{12} & k_{22} & k_{32} & \ddots & \ddots & \vdots \\
    0 & k_{12} & k_{22} & \ddots & \ddots & 0 \\
    0 & \ddots & \ddots & \ddots & k_{32} & 0 \\
    \vdots & \ddots & \ddots & k_{12} & k_{22} & k_{32} \\
    0 & \cdots & 0 & 0 & k_{12} & k_{22}
  \end{array}
\right),\quad
A_{-1}=\left(
  \begin{array}{cccccc}
    k_{23} & k_{33} & 0 & 0 & \cdots & 0 \\
    k_{13} & k_{23} & k_{33} & \ddots & \ddots & \vdots \\
    0 & k_{13} & k_{23} & \ddots & \ddots & 0 \\
    0 & \ddots & \ddots & \ddots & k_{33} & 0 \\
    \vdots & \ddots & \ddots & k_{13} & k_{23} & k_{33} \\
    0 & \cdots & 0 & 0 & k_{13} & k_{23}
  \end{array}
\right),
\end{eqnarray*}
\begin{eqnarray*}
A_{1}=\left(
  \begin{array}{cccccc}
    k_{21} & k_{31} & 0 & 0 & \cdots & 0 \\
    k_{11} & k_{21} & k_{31} & \ddots & \ddots & \vdots \\
    0 & k_{11} & k_{21} & \ddots & \ddots & 0 \\
    0 & \ddots & \ddots & \ddots & k_{31} & 0 \\
    \vdots & \ddots & \ddots & k_{11} & k_{21} & k_{31} \\
    0 & \cdots & 0 & 0 & k_{11} & k_{21}
  \end{array}
\right).
\end{eqnarray*}
In this case, the transformation matrix $M$ corresponding to the convolutional kernel $K$  is a $N^2\times N^2$ doubly  block banded Toeplitz matrix, i.e., a block banded Toeplitz matrix with its blocks are banded Toeplitz matrices. For the details about Toeplitz matrices, please see references  \cite{chan2007,jin2002}. We will let $n=N^2$ and use $\mathcal{T}$ to denote the set of all matrices like $M$ in \eqref{conv0}, i.e., doubly  block banded Toeplitz matrices with the fixed bandth.

For a matrix $M\in\mathcal{T}$,  The value of $K_{p,q}$ will appear in different  $(i,j)$ indexes. We use $\Omega$ to denote this index set, to which each $(i,j)$ index corresponding to $K_{p,q}$ belongs. That is to say,  we have  $m_{ij}=K_{p,q}$ for each $(i,j)\in\Omega$  and  $m_{ij}\neq K_{p,q}$ for each $(i,j)$ that doesn't satisfy $(i,j)\in\Omega$.
\subsection{Regularization 1 to let the Frobeniu norm of $M$ be smaller}
We will use $\frac{1}{2}\|M\|_F^2$ as the penalty function to regularize the convolutional kernel $K$, and calculate $\partial\frac{1}{2} \|M\|_F^2/\partial K_{p,q}$.
The following lemma is easy but useful in the following derivation.
\begin{lemma}\label{lem1}
The partial derivative of square of Frobenius norm of $A\in\mathbb{R}^{n \times n}$ with respect to each entry $a_{ij}$ is $\partial \|A\|_F^2 /\partial a_{ij}=2a_{ij}$.
\end{lemma}

For a matrix $M\in\mathcal{T}$,  The value of $K_{p,q}$ will appear in different  $(i,j)$ indexes. We use $\Omega$ to denote this index set, to which each $(i,j)$ index corresponding to $K_{p,q}$ belongs. That is to say,  we have  $m_{ij}=K_{p,q}$ for each $(i,j)\in\Omega$  and  $m_{ij}\neq K_{p,q}$ for each $(i,j)$ that doesn't satisfy $(i,j)\in\Omega$. The chain rule formula about the derivative tells us that, if we want to calculate $\partial \|M\|_F^2/\partial K_{p,q}$, we should calculate  $\partial \|M\|_F^2/\partial m_{ij}$ for all $(i,j)\in\Omega$ and take the sum, i.e.,

\begin{eqnarray}\label{derivative2}
  \nonumber \frac{1}{2} \frac{\partial \|M\|_F^2}{\partial K_{p,q}}&=& \frac{1}{2}\sum_{(i,j)\in\Omega} \frac{\partial \|M\|_F^2}{\partial m_{ij}}\\
 \nonumber &=& \sum_{(i,j)\in\Omega}m_{ij}.
\end{eqnarray}

We summarize the above results as the following theorem.

\begin{theorem}\label{theo}
Assume  $M\in \mathbb{R}^{n\times n}$ is the doubly  block banded Toeplitz matrix corresponding to the one channel convolution kernel $K\in\mathbb{R}^{k\times k}$. If $\Omega$ is the set of all indexes $(i,j)$ such that $m_{ij}=K_{p,q}$,  we have
\begin{equation}\label{derivative3}
    \frac{1}{2}\frac{\partial \|M\|_F^2}{\partial K_{p,q}}= \sum_{(i,j)\in\Omega}m_{ij}.
\end{equation}
\end{theorem}

Theorem~\ref{theo} provides new insight about how to regularize a convolutional kernel $K$ such that singular values of the corresponding transformation matrix are small. We can use the formula \eqref{derivative3} to carry out the gradient type methods for $\|M\|_F^2$.
\subsection{Regularization 2 to let the smallest singular value of $M$ be larger}
To compute the gradient, we need the following classical result on the first order perturbation expansion about a simple singular value; see \cite{stewart}.
\begin{lemma}\label{lem12}
Let $\sigma$ be a simple singular value of $A = [a_{ij}] \in\mathbb{R}^{m \times m}$
 ($n\geq p$) with normalized left and right singular vectors $u$ and $v$.  Then $\partial \sigma/\partial a_{ij}$  is $u(i)v(j)$, where $u(i)$ is the $i$-th entry of vector $u$ and $v(j)$ is the $j$-th entry of vector $v$.
\end{lemma}

We use the chain rule to get the following theorem.

\begin{theorem}\label{theo1}
Assume the smallest singular value of $M$, which is denoted by $\sigma_{min}(M)$, is simple and positive, where $M\in \mathbb{R}^{n\times n}$ is the doubly  block banded Toeplitz matrix corresponding to the one channel convolution kernel $K\in\mathbb{R}^{k\times k}$. Assume $u$ and $v$  are normalized left and right singular vectors of $M$ associated with $\sigma_{min}(M)$. If $\Omega$ is the set of all indexes $(i,j)$ such that $m_{ij}=K(c,d)$,  we have
\begin{equation}\label{derivative32}
    \partial \sigma_{min}(M)/\partial K(c,d)= \sum_{(i,j)\in\Omega}u(i)v(j).
\end{equation}
\end{theorem}
We can use the formula \eqref{derivative32} to carry out the gradient type methods to let the smallest singular value of $M$ be larger.

\subsection{Regularization 3 to let the  singular values of $M$ be neither large nor small}
We can combine Theorem~\ref{theo} and Theorem~\ref{theo1} to let the  singular values of $M$ be neither large nor small.
As we know, $\|M\|_F^2$ is the squares sum  of  all singular values of $M$. If $M$ is  $n\times n$,  $\|M\|_F^2$ is the squares sum of $n$ singular values. We may choose $-n\sigma_{min}(M)+\frac{1}{2} \|M\|_F^2$ as the regularization term to let the singular values of $M$ be neither large nor small. Thus we have the following theorem.

\begin{theorem}\label{theo1}
Assume the smallest singular value of $M$, which is denoted by $\sigma_{min}(M)$, is simple and positive, where $M\in \mathbb{R}^{n\times n}$ is the doubly  block banded Toeplitz matrix corresponding to the one channel convolution kernel $K\in\mathbb{R}^{k\times k}$. Assume $u$ and $v$  are normalized left and right singular vectors of $M$ associated with $\sigma_{min}(M)$. If $\Omega$ is the set of all indexes $(i,j)$ such that $m_{ij}=K(c,d)$,  we have
\begin{equation}\label{derivative32}
    \partial (\frac{1}{2} \|M\|_F^2-n\sigma_{min}(M))/\partial K(c,d)= \sum_{(i,j)\in\Omega}(m_{ij}-nu(i)v(j)).
\end{equation}
\end{theorem}

\section{The penalty function and the gradient for multi-channel convolution}\label{sec:multi}
In this section we consider the case of multi-channel convolution. First we show the transformation matrix corresponding to multi-channel convolution.
At each convolutional layer,  we have convolution kernel $K\in \mathbb{R}^{k\times k\times g  \times h}$ and the input $X\in \mathbb{R}^{N\times N\times g}
$; element $X_{i,j,d}$ is the value of the input unit within channel $d$ at row $i$ and column $j$. Each entry of  the output $Y\in \mathbb{R}^{N \times N \times h}$  is produced by
\begin{equation*}
    Y_{r,s,c}= (K*X)_{r,s,c}=\sum_{d\in \{1,\cdots,g\}} \sum_{p\in \{1,\cdots,k\} }\sum_{q\in \{1,\cdots,k\}} X_{r-m+p,s-m+q,d}K_{p,q,d,c},
\end{equation*}
where $X_{i,j,d}=0$ if $i\leq 0$ or $i> N$, or $j\leq 0$ or $j> N$.
By inspection, $vec(Y)=Mvec(X)$, where $M$ is as follows
\begin{eqnarray}\label{conv2}
M=\left(\begin{array}{cccc}
M_{(1)(1)} & M_{(1)(2)} & \cdots& M_{(1)(g)} \\
M_{(2)(1)} & M_{(2)(2)} & \cdots & M_{(2)(g)} \\
\vdots &\vdots &\cdots  &\vdots \\
M_{(h)(1)}& M_{(h)(2)}& \cdots &M_{(h)(g)}
\end{array}\right),
\end{eqnarray}
and each $M_{(c)(d)}\in \mathcal{T}$, i.e., $M_{(c)(d)}$ is a $N^2\times N^2$ doubly  block banded Toeplitz matrix corresponding to the portion $K_{:,:,d,c}$ of $K$ that concerns the effect of the $d$-th input channel on the $c$-th output channel.

Similar as the proof in Section~\ref{sec:one},  we have the following theorem.
\begin{theorem}\label{theo2}
Assume  $M$  is the structured matrix corresponding to the multi-channel convolution kernel $K\in\mathbb{R}^{k\times k\times g \times h}$ as defined in (\ref{conv2}). Given $(p,q,z,y)$, if $\Omega_{p,q,z,y}$ is the set of all indexes $(i,j)$ such that $m_{ij}=k_{p,q,z,y}$,  we have
\begin{equation}\label{derivative4}
    \frac{1}{2}\frac{\partial \|M\|_F^2}{\partial K_{p,q,z,y}}= \sum_{(i,j)\in\Omega_{p,q,z,y}}m_{ij}.
\end{equation}
\end{theorem}
Then the  gradient descent algorithm for the penalty function $\|M\|_F^2$ can be devised, where the number of channels maybe more than one.
\begin{theorem}\label{theo21}
Assume  $M$  is the structured matrix corresponding to the multi-channel convolution kernel $K\in\mathbb{R}^{k\times k\times g \times h}$ as defined in (\ref{conv2}). Given $(p,q,z,y)$, if $\Omega_{p,q,z,y}$ is the set of all indexes $(i,j)$ such that $m_{ij}=k_{p,q,z,y}$,  we have

\begin{equation}\label{derivative41}
    \partial \sigma_{min}(M)/\partial K_{p,q,z,y}=\sum_{(i,j)\in\Omega_{p,q,z,y}}u(i)v(j).
\end{equation}
\end{theorem}

We present the detailed gradient descent algorithm for the three different penalty functions, where in Algorithm\ref{alg3}, $min(g,h)$ denotes the smaller one of $g$ and $h$.

\begin{algorithm}\label{alg1}
\noindent \textbf{Gradient Descent for ${\cal R}_\alpha (K) =\frac{1}{2} \|M\|_F^2$}
\begin{tabbing}
aaaaa \= bbbb\= \kill
1. \> Input: an initial kernel $K\in \mathbb{R}^{k\times k\times g  \times h}$, input size $N\times N\times g$   and learning rate $\lambda$.\\
2. \>While not converged:\\
3. \>\>Compute $G= [\frac{\frac{1}{2} \|M\|_F^2}{\partial k_{p,q,z,y}}]_{p,q,z,y=1}^{k,k,g,h} $, by \eqref{derivative4}; \\
4. \>\> Update $K=K-\lambda G$;\\
5. \>End
\end{tabbing}
\end{algorithm}

\begin{algorithm}\label{alg2}
\noindent \textbf{Gradient Descent for ${\cal R}_\alpha (K) =-\sigma_{min}(M)$}
\begin{tabbing}
aaaaa \= bbbb\= \kill
1. \> Input: an initial kernel $K\in \mathbb{R}^{k\times k\times g  \times h}$, input size $N\times N\times g$   and learning rate $\lambda$.\\
2. \>While not converged:\\
3. \>\>Compute $G= [\frac{-\sigma_{min}(M)}{\partial k_{p,q,z,y}}]_{p,q,z,y=1}^{k,k,g,h} $, by \eqref{derivative41}; \\
4. \>\> Update $K=K-\lambda G$;\\
5. \>End
\end{tabbing}
\end{algorithm}

\begin{algorithm}\label{alg3}
\noindent \textbf{Gradient Descent for ${\cal R}_\alpha (K) =\frac{1}{2} \|M\|_F^2-min(g,h)N^2\sigma_{min}(M)$}
\begin{tabbing}
aaaaa \= bbbb\= \kill
1. \> Input: an initial kernel $K\in \mathbb{R}^{k\times k\times g  \times h}$, input size $N\times N\times g$   and learning rate $\lambda$.\\
2. \>While not converged:\\
3. \>\>Compute $G= [\frac{\frac{1}{2} \|M\|_F^2-min(g,h)N^2\sigma_{min}(M)}{\partial k_{p,q,z,y}}]_{p,q,z,y=1}^{k,k,g,h} $, by \eqref{derivative4} and \eqref{derivative41}; \\
4. \>\> Update $K=K-\lambda G$;\\
5. \>End
\end{tabbing}
\end{algorithm}

\section{Numerical experiments}\label{sec:numer}
\begin{figure}[h]
  \centering
  \includegraphics[width=1.00\textwidth]{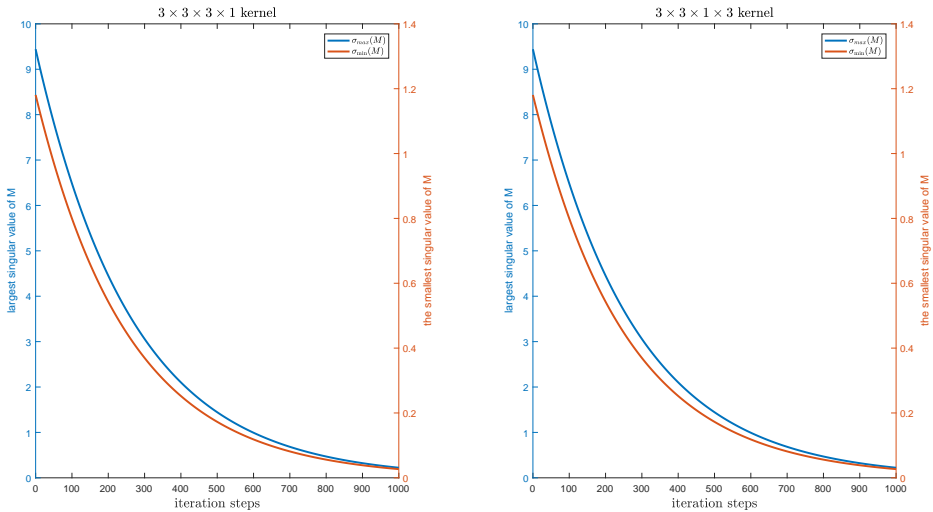}
\caption{\text{\small{Changes of $\sigma_{max}(M)$ and $\sigma_{min}(M)$ for different kernel sizes}}}
\label{figexplode}
  \end{figure}

\begin{figure}[h]
  \centering
  \includegraphics[width=1.00\textwidth]{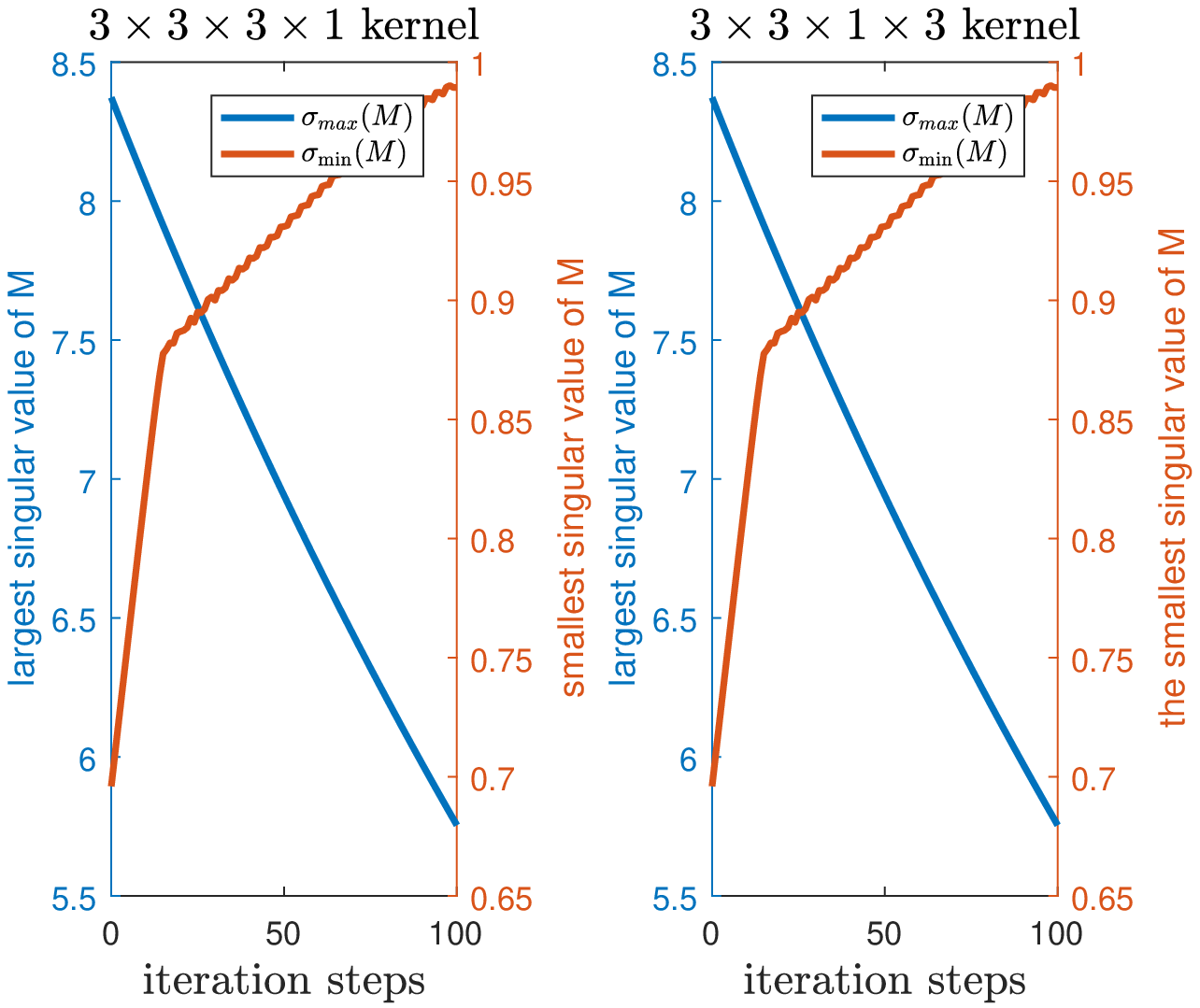}
\caption{\text{\small{Changes of $\sigma_{max}(M)$ and $\sigma_{min}(M)$  for different kernel sizes}}}
\label{figunstable}
  \end{figure}

The numerical  tests were performed on a laptop (3.0 Ghz and 16G Memory) with MATLAB R2016b.
We use $M$ to denote the transformation matrix corresponding to the convolutional kernel. The largest singular value and smallest singular value of $M$ (denoted as ``$\sigma_{max}(M)$ and $\sigma_{min}(M)$), the iteration steps (denoted as ``iter") are demonstrated to show the effectiveness of our method.
Numerical experiments are implemented on extensive test problems.
In this paper we present the numerical results for some random generated multi-channel convolution kernels, where $K$ is generated by the following command
\begin{quote}
    rand('state',1);\\
    $K= rand(k,k,g,h);$
\end{quote}
We consider kernels of different sizes with $3\times 3$ filters, namely $K\in \mathbb{R}^{3\times 3\times g  \times h}$ for various values of $g,h$.  For each kernel, we use $20\times 20\times g$ as the size of input data matrix.  We then minimize the three different penalty functions using Algorithm \ref{alg1}, \ref{alg2} and \ref{alg3} respectively. We show the effects of  changing the singular values of $M$. We present in Figures the results for $3\times 3\times 3  \times 1$ and $3\times 3\times 1  \times 3$ kernels. In the figures \ref{figexplode} and \ref{figunstable}, we show the convergence of $\sigma_{max}(M)$  on the left axis scale and $\sigma_{min}(M)$  on the right axis scale.

\begin{figure}[h]
  \centering
  \includegraphics[width=1.00\textwidth]{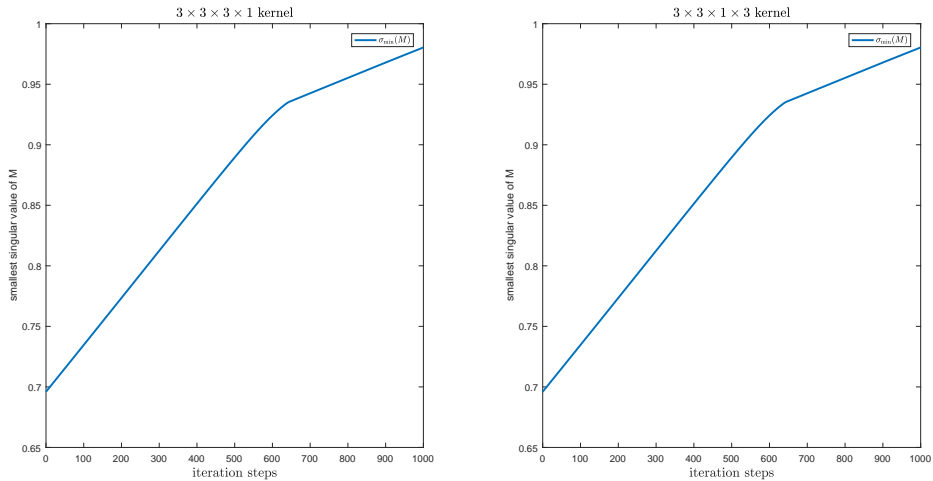}
\caption{\text{\small{Changes of $\sigma_{min}(M)$ for different kernel sizes}}}
\label{figvanish}
 \end{figure}

 Numerical experiments are done on other random generated examples, including  random kernels with each entry uniformly distributed on $[0, 1]$.
The convergence figures of $\sigma_{max}(M)$ and $\sigma_{min}(M)$ are similar with the subfigures presented in the paper.

The efficiency  of each method, i.e., the needed iteration steps to let $\sigma_{max}(M)$ and $\sigma_{min}(M)$ be bounded in a satisfying interval, is related with the step size $\lambda$. In our numerical experiments, for Algorithms \ref{alg1} and \ref{alg3} we  use  the  step size $\lambda=1e-5$ while for Algorithms \ref{alg2} we use the step size $\lambda=1e-4$.
We can't definitely tell how to choose the optimal step size currently.

\section{Conclusions}\label{sec:conclu}
In this paper, we provide a new regularization method to  regularize the weights of convolutional layers in deep neural networks. We give new regularization terms about convolutional kernels to change the singular values of the corresponding structured transformation matrices. We propose  gradient decent algorithms for the regularization terms. This method is shown to be effective.

In future, we will  continue to  devise other forms of penalty functions for convolutional kernels to constrain the singular values of  corresponding  transformation matrices.
\section{Acknowledgements}


\begin{thebibliography}{99}
\bibitem{brock2017}
Andrew Brock, Theodore Lim, James M Ritchie, and Nick Weston. Neural photo editing with introspective
adversarial networks. In ICLR, 2017.
\bibitem{chan2007}
 R. Chan and X. Jin, An Introduction to Iterative Toeplitz Solvers, SIAM, Philadelphia, 2007.
 \bibitem{cisse}
 Moustapha Cisse, Piotr Bojanowski, Edouard Grave, Yann Dauphin, Nicolas Usunier.
 Parseval Networks: Improving Robustness to Adversarial Examples. In ICML, 2017.
 \bibitem{dumoulin2018}
 Vincent Dumoulin, Francesco Visin. A guide to convolution arithmetic for deep learning. ArXiv, 2018.
 \bibitem{golub2012}
G.-H. Golub and  C.-F. Van Loan, Matrix computations, Johns Hopkins University Press, Baltimore, 2012.
\bibitem{goodfellow2013}
 I. J. Goodfellow, J. Shlens, and C. Szegedy. Explaining and harnessing adversarial examples. In ICLR,
2015.
\bibitem{guo2019}
P. Guo, Q. Ye. On Regularization of Convolutional Kernels in Neural Networks, ArXiv 2019.
\bibitem{guo20192}
P. Guo. A Frobenius norm regularization method for convolutional kernels to avoid unstable gradient problem, ArXiv 2019.
\bibitem{hochreiter2001}
S. Hochreiter, Y. Bengio, P. Frasconi, J. Schmidhuber, et al. Gradient flow in recurrent nets: the difficulty of
learning long-term dependencies, In Field Guide to Dynamical Recurrent Networks, IEEE Press, 2001.
\bibitem{jin2002}
 X. Jin, Developments and Applications of Block Toeplitz Iterative Solvers, Science Press, Beijing, 2002.

%\bibitem{keskar}
%N. S. Keskar, D. Mudigere, J. Nocedal, M. Smelyanskiy, and P. T. P. Tang. On large-batch training for
%deep learning - generalization gap and sharp minima. In ICLR, 2017.
%\bibitem{lecun2012}
%Y. A. LeCun, L. Bottou, G. B. Orr, and K.-R. Muller. Efficient backprop. In Neural networks: Tricks of
%the trade, pages 9-48. Springer, 2012.
\bibitem{kova2008}
Kova$\breve{c}$evi$\acute{c}$, Jelena and Chebira, Amina. An introduction to frames, Now Publishers Inc, Boston,
2008.
\bibitem{miyato2018}
Takeru Miyato, Toshiki Kataoka, Masanori Koyama, Yuichi Yoshida. Spectral Normalization for Generative Adversarial Networks. In ICLR, 2018.
\bibitem{sedghi2018}
Hanie Sedghi, Vineet Gupta and Philip M. Long. The Singular Values of Convolutional Layers. In ICLR, 2019.
\bibitem{stewart}
G. W. Stewart. Matrix Algorithms: Volume II. Eigensystems, SIAM, 2001.
\bibitem{szegedy2014}
C. Szegedy, W. Zaremba, I. Sutskever, J. Bruna, D. Erhan, I. J. Goodfellow, and R. Fergus. Intriguing
properties of neural networks. In ICLR, 2014.
\bibitem{tsuzuku2018}
Y. Tsuzuku, I. Sato,  and M. Sugiyama.
Lipschitz-Margin Training: Scalable Certification of Perturbation Invariance for Deep Neural Networks.
In NIPS, 2018.
\bibitem{Orthberkely}
Jiayun Wang, Yubei Chen, Rudrasis Chakraborty, and Stella X. Yu. Orthogonal Convolutional Neural Networks. ArXiv 2019
\bibitem{zhang}
C. Zhang, S. Bengio, M. Hardt, B. Recht, and O. Vinyals. Understanding deep learning requires rethinking generalization. In ICLR, 2017.
\bibitem{yoshida}
Yuichi Yoshida, Takeru Miyato. Spectral Norm Regularization for Improving the Generalizability of Deep Learning, ArXiv 2017.




\end{thebibliography}
\end{document}